# ARPA: Armenian Paraphrase Detection Corpus and Models

Arthur Malajyan[1], Karen Avetisyan[2], Tsolak Ghukasyan[3]
Ivannikov Laboratory for System Programming at Russian-Armenian University,
Yerevan, Armenia
{[1]malajyanarthur, [2]karavet, [3]tsggukasyan}@ispras.ru

*Abstract*— In this work, we employ a semi-automatic method based on back translation to generate a sentential paraphrase corpus for the Armenian language. The initial collection of sentences is translated from Armenian to English and back twice, resulting in pairs of lexically distant but semantically similar sentences. The generated paraphrases are then manually reviewed and annotated. Using the method train and test datasets are created, containing 2360 paraphrases in total. In addition, the datasets are used to train and evaluate BERT-based models for detecting paraphrase in Armenian, achieving results comparable to the state-of-the-art of other languages.

## I. INTRODUCTION

Paraphrase detection is the task of verifying that a pair of text fragments are semantically identical. It has valuable applications in various natural language processing tasks, particularly plagiarism detection and text summarization. Since there is no formal definition of paraphrase, researchers relied on data-driven methods when approaching the detection task. For that purpose, the existence of paraphrase-annotated corpora is essential.

Review of literature revealed that there have been no publicly available paraphrase detection resources for the Armenian language. This work is devoted to creating a corpus of Armenian sentential paraphrases for training and evaluation of paraphrase detection models.

The creation of such corpora poses several challenges, and obtaining diverse paraphrases with semantically similar but lexically distant pairs of sentences is one of them. There are several approaches to creating paraphrase datasets, which can be grouped into (i) monolingual paraphrase by experts, (ii) semi-automatic with post-editing by experts, (iii) fully automatic. Federmann et al. carried out a study to compare these techniques, and concluded that using machine translation for paraphrase generation is a well-performing approach that, compared to human experts, is significantly less costly and leads to more diverse examples [1]. At the same time, they recommended post-editing translation-generated paraphrases by human experts to improve their fluency and adequacy. Therefore, we adopted a similar approach and used back translation and subsequent manual review to generate paraphrases of Armenian sentences. Our approach differed from Federmann et al., 2019 in that we repeated the back translation step twice to achieve increased diversity, and then during post-editing stage human experts only verified the fluency and adequacy of the generated examples to exclude incorrect sentences from the datasets.

Apart from the generation method, this work mainly followed the recommendations from Dolan et. al. [2] and Pivovarova et. al. [3], and used them as a point of reference. Dolan et. al. describe the creation of MSRP corpus, consisting of 5801 news cluster pairs and used for evaluation of English paraphrase detection models. Methods that we use in our work for extraction task are also used in MSRP and are described in [4] and [5]. Pivovarova et. al. introduced the ParaPhraser corpus for the Russian language, consisting of headlines of news articles and based on the work of Wubben et al. [6], where similarity metric is used for paraphrase candidate extraction.

In addition to datasets, we also developed paraphrase detection models for the Armenian language. Taking into account the fact that machine learning models, BERT specifically, have shown state-of-the-art results on paraphrase detection tasks over the last few years [7][8], we decided to employ Multilingual BERT to fine-tune for paraphrase detection. Multilingual BERT supports Armenian, and the decision to use it is also explained by the lack of monolingual Armenian BERT, training of which from scratch would be challenging because of the cost and the lack of big textual corpora. The datasets and models developed in this work are publicly available on GitHub[1].

### A. Related Work

Wieting et al. also used neural machine translation to generate sentential paraphrases via back-translation of bilingual sentence pairs for the training of sentence embeddings [9]. Apart from machine translation, other paraphrase generation techniques have been explored, such as rule-based [10], reinforcement learning-based [11], seq2seq [12][13][14][15] and others.

A lot of previous research focused on finding naturally-occurring sentential paraphrases [16][17][18]. There were attempts to base corpora on movie subtitles like Opusparcus multilingual corpus for six languages: German, English, Finnish, French, Russian, Swedish [19]. The extraction stage of TMUP corpus is similar to ours and is based on two different translation mechanisms Google PBMT and NMT [20][21]. Some languages like Arabic have specific transformation rules and paraphrasing mechanism can be done by using them, which is shown in [22].

## II. DATASETS

The corpus was created to resolve the problem of training and evaluating sentential paraphrase detection models. The first subsection describes the selection process of the initial set of sentences. The second subsection describes the method based on back translation for generating the paraphrases. The final subsection is dedicated to the manual annotation of the obtained pairs.

---

[1] https://github.com/ivannikov-lab/arpa-paraphrase-corpus

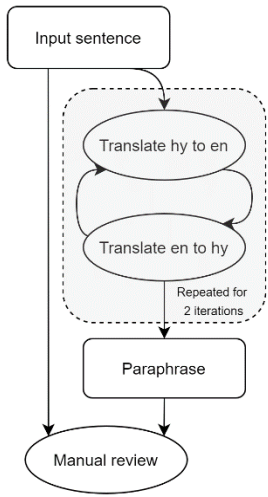

Fig. 1a. The paraphrase generation scheme via translation to English (en) and back to Armenian (hy).

Fig. 1b. Examples of generated paraphrases (non-overlapping words are underlined; below each sentence expert translation is given in grey).

| Input sentence: | Paraphrase: |
|---|---|
| Կոռուպցիան չարիք են համարում բոլորը՝ չինովնիկից մինչև բանվոր: Corruption is considered evil by everyone, from chinovnik to worker. | Կոռուպցիան բոլորի համար չարիք է համարվում՝ պաշտոնյաներից մինչև աշխատակիցներ: Corruption is considered bad for everyone, from officials to employees. |
| Քաղաքացիներից մեկն էլ «Հետք»-ին ուղարկած նամակում նույնիսկ նշել էր, որ Հայաստանում Լեհաստանի դիվանագիտական ներկայության վերանայման կարիք ունի: In a letter to Hetq, one of the citizens even mentioned that the Polish diplomatic presence in Armenia needs to be reconsidered. | Մի քաղաքացի նույնիսկ «Հետքին» գրեց, որ Լեհաստանի դիվանագիտական ներկայությունը Հայաստանում պետք է վերանայվի: One citizen even wrote to Hetq that Poland's diplomatic presence in Armenia should be reconsidered. |
| Կարինեն սովորել է Նոյեմբերյանի պետական քոլեջի հաշվապահության բաժնում, վերջերս ստացել է իր դիպլոմը: Karine studied at Noyemberyan State College's department of accounting, recently received her diploma. | Կարինեն սովորում էր հաշվապահություն Նոյեմբերյանի պետական քոլեջում և վերջերս դիպլոմ ստացավ: Karine was studying accounting at Noyemberyan State College and has received a diploma recently. |

*A. Sentence Selection*

For this task were used news texts consisting of articles written in the last 10 years crawled from Hetq (12,122 articles) and Panarmenian (12,497 articles) news websites. The set covers texts about different topics: politics, sports, economy, etc.

Upon receiving the initial set of sentences, it appeared that some of them were poor for inclusion in paraphrase corpus thus had to be filtered as follows. At first, sentences that contained information about the page or the section they were pointing to or they were at, contained meaningless information and were removed (e.g. Հայաստանի Հանրապետության արտաքին առևտուրը 2007 թվականին, Վիճակագրական Ժողովածու, Երևան , 2008 , էջ 9697: // Foreign Trade of the Republic of Armenia 2007, Statistical Collection, Yerevan, 2008, page 9697).

For some texts, sentence boundaries were detected incorrectly and we ended up with either too long or too short sentences. To prevent this kind of pairs from appearing in our final set we removed all sentences containing fewer than 6 tokens and more than 22 tokens (not counting stopwords). Furthermore, if a sentence contained three or more identical words in a row, it was also removed.

*B. Sentence Pair Generation Using Back Translation*

The method used for generating semantically similar sentences is based on Armenian to English and vice versa sentence translation. Google Translate is one of the few available translators for the Armenia language, demonstrates relatively high accuracy, and therefore was selected for translation. We also considered translation from Armenian to Russian, however the translation accuracy was visibly worse than for English.

In this part, the sentences selected in previous section were taken and translated back and forth (Figure 1a). The back translation process was repeated twice. We observed that after one iteration the generated sentences still retained a high level of lexical and morphosyntactic similarity, while 3 or more iterations led to higher proportion of erroneous translations.

Translated sentences that contained symbols from two different scripts in one word were also removed from the set (e.g. genocideաբանություն). The original sentence and its translation were considered as a sentence pair in our set.

With a perfect translator, utilizing this method would allow to obtain as many pairs as desired. However, Google Translate obviously makes translation mistakes, some of which result in meaningless translations or translations that are no longer the paraphrase of the original sentence. Therefore, we had to annotate the obtained data to separate paraphrase pairs from non-paraphrase pairs or even from the pairs which contained wrong translation.

From generated sentence pairs we manually filtered those that contained translations that were syntactically or semantically incorrect, partly translated (i.e. consisting of predominantly foreign words), or contained multiple sentences. This way, 1450 out of 4405 reviewed sentence pairs (roughly the third) were removed. The remaining 2955 pairs were further examined manually, as described in the following section.

*C. Annotation*

After filtering erroneous pairs, we proceeded with annotating the rest of the pairs, mainly relying on annotators' judgment to decide whether it is paraphrase or not. To increase the agreement in the dataset, the annotators were given a guideline, roughly following the 2012 SemEval's Semantic Textual Similarity degrees to differentiate paraphrase from non-paraphrase [23]. Pairs with similarity degrees 5 ("Completely equivalent") and 4 ('Mostly equivalent, but some unimportant details differ") were annotated as paraphrase, and degrees 0 ("On different topics") to 3 ("Roughly equivalent, but some important information differs/missing") were annotated as non-paraphrase.

Each annotator was given a list of specific examples as to what should not be considered as paraphrase, including near paraphrases such as:

I. partially overlapping sentences, e.g.:

| Այսօր 100%-ով վերականգնվել է էլեկտրամատակարարումը - հայտարարել է նախարար Խորխե Ռոդրիգեսը: The power supply has been fully restored today, said Minister Jorge Rodriguez. | Այսօր էլեկտրոէներգիայի վերականգնվել է 100% -ի չափով: Today, electricity is 100% restored. |

II. pairs with strictly unidirectional entailment, e.g.:

*Բայց, միևնույն ժամանակ, դա պարտավորեցնում է, որ ես ավելի շատ պարապեմ:*
But at the same time, it obliges me to train even more.

*Բայց, միևնույն ժամանակ, դա ինձ ստիպում է ավելին անել:*
But at the same time, it forces me to do more.

III. pairs with similar/identical context but referring to different entities, e.g.:

*Լիանան ռուսաստանցի երգչուհի է, որը կատարում է էլեկտրոնային և էթնիկ ոճի երաժշտություն:*
Linda is a Russian singer who performs electronic ethnic music.

*Սվետլանան ռուս երգչուհի է, ով նվագում է էլեկտրոնային և էթնիկ ոճի երաժշտություն:*
Svetlana is a Russian singer who plays electronic ethnic music.

*Հիշեցնենք, որ նա մեղադրվում է ՀՀ քրեական օրենսգրքի 300.1-րդ հոդվածի 1-ին մասով:*
It should be reminded that he is charged with Part 1 of Article 300.1 of the RA Criminal Code.

*Նրան մեղադրանք է առաջադրվել ՀՀ քրեական օրենսգրքի 311-րդ հոդվածի 1-ին մասով:*
He was charged with Article 311, Part 1 of the RA Criminal Code.

The set of sentence pairs was divided into 2 subsets (1573 for training and 1382 for testing). Using the described guide, train set was manually reviewed by 1 annotator. After manual examination, 1339 out of 1573 train examples were considered as "paraphrase" (85%). For test set, each pair was reviewed by at least 2 annotators. Disagreements were resolved by a third annotator. The inter-annotator agreement, measured using Cohen's Kappa, varied from 0.55 to 0.65 between annotator pairs, which is comparable to the scores for MRPC and Paraphraser datasets. After review, 1021 test pairs were labeled as paraphrase out of total 1382 (74%). Overall, 80% of automatically generated sentence pairs were confirmed as paraphrase after manual review. The rest of the pairs that were deemed non-paraphrase, still had high semantic similarity and were roughly equivalent, but with some important details differing (Figure 2).

Fig. 2. Examples of translations that were labelled as non-paraphrase.

| Original sentence | Translation |
|---|---|
| *Եթե նման ցանկություն ունեն, ազատվելու են գնան, անփոխարինելի մարդ չկա՝ ինձնից սկսված»,- վստահեցրեց ՀՀ վարչապետը:* "If they have such a wish, they will be released, no person is irreplacable, starting with me," RA Prime Minister assured. | *Եթե նրանք ունենան նման ցանկություն, նրանք կազատվեն, ինձանից անփոխարինելի մարդ չկա», - վստահեցրեց վարչապետը:* "If they have such a wish, they will be released, no person is more irreplacable than me," the Prime Minister assured. |
| *Այլ կերպ ասած՝ ինչից շատ ունենք, դա էլ ցույց ենք տալիս:* In other words, we show that which we have most. | *Այլ կերպ ասած, մենք ցույց ենք տալիս ավելին, քան ունենք:* In other words, we show more than what we have. |
| *Շիրակի մարզում տարիներ շարունակ պատկերը մնացել է նույնը:* In Shirak region, the picture [situation] has remained the same for years. | *Շիրակում նկարը տարիներ շարունակ մնացել է նույնը:* In Shirak, the picture [image] has remained the same for years. |
| *Այն հեղինակել է «Ազատություն Լևոն Հայրապետյանին» քաղաքացիական նախաձեռնությունը:* It was authored by the "Freedom to Levon Hayrapetyan" civil initiative. | *Այն հեղինակել է «Ազատություն» - ը՝ Լևոն Հայրապետյանի քաղաքացիական նախաձեռնության համար:* It was authored by "Azatutyun" for Leon Hayrapetyan's civil initiative. |
| *Նրանց տեղը գրաղեցրել է ֆրանսահայ Քրիստոնյա Զադիկյանը:* They were replaced by French-Armenian Christian [given name] Zadikyan. | *Նրանց տեղը գրավել ֆրանսահայ քրիստոնյա Զադիկյանը:* They were replaced by French-Armenian Christian [follower of Christianity] Zadikyan. |

Following [1], we also verified the diversity of generated paraphrases by computing the average number of word-level edits between the source sentence and its paraphrase. When compared to the diversity scores of MRPC and ParaPhraser datasets, our paraphrases (named ARPA) demonstrated greater level of diversity (Table I). It should be noted that the diversity score did not count punctuation and stop-words, to better reflect meaningful changes.

TABLE I. Comparison of paraphrase diversity level in English, Russian, and Armenian datasets.

| Dataset | Paraphrase diversity | |
|---|---|---|
| | *Train set* | *Test set* |
| MRPC | 6.79 | 7.01 |
| ParaPhraser.ru | 5.02 | 5.51 |
| **ARPA** | **8.70** | **8.66** |

### D. Negative Examples

Furthermore, we appended the obtained sets with automatically generated negative pairs. For train set, 2660 non-paraphrase sentence pairs were generated. Half of those pairs were consecutive sentences, which we assumed would have some overlap. The other half were generated by taking two random sentences from texts. Similarly, we also added a relatively small number of negative pairs to the test set (150 consecutive and 150 random), for better representation of the sentence space. When compared to Russian and English datasets (Table II), our test set has a comparable size and contains a similar number of paraphrases.

TABLE II. Label distributions for Russian, Armenian and English sets.

| Dataset | Paraphrase | | Non-paraphrase | | Total |
|---|---|---|---|---|---|
| | *Examples* | *Average Jaccard similarity* | *Examples* | *Average Jaccard similarity* | |
| *Test* | | | | | |
| MRPC | 1147 | 0.438 | 578 | 0.322 | 1725 |
| ParaPhraser.ru | 1137 | 0.317 | 762 | 0.169 | 1899 |
| **ARPA** | 1021 | 0.327 | 661 | 0.172 | 1682 |
| *Train* | | | | | |
| MRPC | 2753 | 0.444 | 1323 | 0.325 | 4076 |
| ParaPhraser.ru | 4255 | 0.306 | 2947 | 0.119 | 7202 |
| **ARPA** | 1339 | 0.320 | 2894 | 0.056 | 4233 |

## III. PARAPHRASE DETECTION

### A. Models

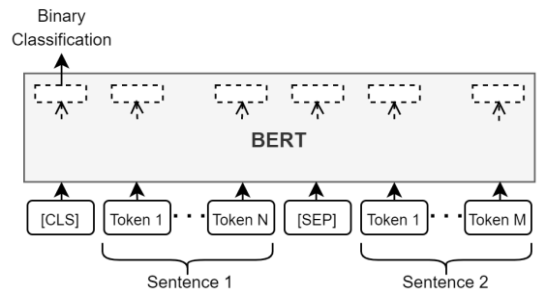

Fig. 3. The paraphrase detection model.

Based on the success of BERT-based models in the task of sentential paraphrase detection, we adopted a similar model (Figure 3). In our experiments, we train the model on the proposed ARPA dataset and compare it with models trained on the translations of English and Russian corpora. In addition, we compare the results with the performance of English and Russian paraphrase detection tools on our test set, by translating the examples to the respective language using Google Translate. The list of explored models is given below:

a. Multilingual BERT, fine-tuned on the following datasets:
   i. MRPC, translated to Armenian,
   ii. ParaPhraser.ru, translated to Armenian,

iii. ARPA dataset, proposed in this work,
iv. All of the training sets above combined.

b. DeepPavlov's RUBERT-based paraphrase identification tool, tested on ARPA google-translated to Russian,
c. BERT-Base trained on MRPC and tested on ARPA google-translated to English.

**Hyperparameters:** When finetuning Multilingual BERT, we used 0.00002 learning rate, 0.5 dropout rate, and 32 batch size. Sequence length was limited to 64 tokens.

*B. Results and Discussion*

The performance results of the described models are given in Table III. Overall, the multilingual BERT models trained on annotated Armenian examples produced the best results. Surprisingly, the results of RUBERT were quite close and even noticeably higher in terms of recall. This suggests that it might be worth exploring dataset generation via back translation to Russian. English BERT model was also able to detect translated paraphrase, however its precision was the worst among all models.

TABLE III. Models' performance on the proposed test set.

| Model | Scores (95% confidence interval) | | | |
|---|---|---|---|---|
| | F1 | Accuracy | Recall | Precision |
| a.i. trMRPC | 0.801 ± 0.014 | 0.699 ± 0.028 | 0.993 ± 0.005 | 0.672 ± 0.021 |
| a.ii. trParahraser | 0.838 ± 0.002 | 0.771 ± 0.002 | 0.977 ± 0.005 | 0.734 ± 0.002 |
| a.iii. ARPA | 0.837 ± 0.003 | 0.775 ± 0.003 | 0.952 ± 0.009 | 0.747 ± 0.002 |
| a.iv. Combined | 0.840 ± 0.002 | 0.776 ± 0.002 | 0.971 ± 0.006 | 0.741 ± 0.001 |
| b. RUBERT | 0.837 | 0.764 | 0.998 | 0.721 |
| c. BERT | 0.779 | 0.656 | 1.0 | 0.638 |

**Performance on near-paraphrase examples:** While labeling sentence pairs we additionally marked near-paraphrases. These were the semantically close examples which were difficult to differentiate from paraphrase. We separately calculated the accuracy of models on these examples (Table IV).

TABLE IV. Accuracy of models on near-paraphrase pairs.

| Model | Accuracy on near-paraphrases |
|---|---|
| a.i. tr-MRPC | 3.00% |
| a.ii. tr-ParaPhraser | 4.17% |
| a.iii. ARPA | 9.05% |
| a.iv. Combined | 4.55% |

The models scored very low on the subset. Multilingual BERT fine-tuned on ARPA train set performed the best, showing only 9.05% accuracy. This is not surprising however, as the examples were hard to label even for human annotators.

TABLE V. The comparison of BERT-based paraphrase detection state-of-the-art models for English, Russian and Armenian languages.

| Dataset | BERT Model | F1 | Accuracy | Recall | Precision |
|---|---|---|---|---|---|
| MRPC | BERT-Base | 88.9 | 83.5 | 99.38 | 80.39 |
| Paraphraser.ru | RUBERT | 87.9 | 84.9 | 91.60 | 84.48 |
| | BERT-Base Multilingual | 83.4 | 79.3 | 86.84 | 80.22 |
| ARPA | BERT-Base Multilingual | 83.7 | 77.5 | 95.20 | 74.70 |

**Comparison with other languages:** The results obtained on ARPA were compared to the results of best BERT-based paraphrase detection models on English and Russian datasets (Table V). Given the substantially smaller train set, we still were able to achieve comparable results in terms of recall. Precision of the trained model was significantly lower, however. Apart from the size of the training set, this potentially could be caused by the quality of multilingual BERT's parameters for the Armenian language. It is worth noting that the best results for English and Russian were obtained using monolingual BERT models.

IV. CONCLUSION

We used back translation to generate a sentential paraphrase corpus for the Armenian language. The generated paraphrases were manually reviewed and annotated, resulting in gold standard train and test datasets containing 2360 paraphrases in total. The datasets were used to train and evaluate BERT-based models for detecting paraphrase in Armenian, establishing a point of reference for future research.

ACKNOWLEDGMENT

The authors thank Denis Turdakov, Yaroslav Nedumov and Kirill Skornyakov for their insightful feedback. The authors also thank Marine Mikilyan and Arman Darbinyan for helping organize the annotation of training examples, as well as the linguistics students at Russian-Armenian University who participated in the annotation.